\documentclass[lettersize,journal]{IEEEtran}
\usepackage{amsmath,amsfonts}
\usepackage{algorithmic}
\usepackage{array}
\usepackage[caption=false,font=normalsize,labelfont=sf,textfont=sf]{subfig}
\usepackage{textcomp}
\usepackage{stfloats}
\usepackage{url}
\usepackage{verbatim}
\usepackage{graphicx}
\usepackage{float}
\usepackage{booktabs}
\usepackage{multirow}
\usepackage{bbding}
\usepackage{diagbox}
\usepackage{amssymb}
\def\BibTeX{{\rm B\kern-.05em{\sc i\kern-.025em b}\kern-.08em
    T\kern-.1667em\lower.7ex\hbox{E}\kern-.125emX}}
\usepackage{balance}
\begin{document}
\title{Remote Sensing Image Segmentation Using Vision Mamba and Multi-Scale Multi-Frequency Feature Fusion}
\author{Yice Cao, Chenchen Liu, Zhenhua Wu, Wenxin Yao, Liu Xiong, Jie Chen, Zhixiang Huang
\thanks{This work was supported in part by the National Natural Science Foundation of China under Grant 62401007, 62201007 and Grant 62001003, in part by the Natural Science Foundation of Anhui Province under Grant 2308085QF199, in part by the Open Project Funds for the Key Laboratory of Space Photoelectric Detection and Perception (Nanjing University of Aeronautics and Astronautics), Ministry of Industry and Information Technology under Grant NJ2024027-4, in part by the Fundamental Research Funds for the Central Universities under Grant NJ2024027. \emph{(Corresponding author: Zhenhua Wu)}

Yice Cao, Chenchen Liu, Wenxin Yao, Liu Xiong, and Zhixiang Huang are with the Key Laboratory of Intelligent Computing and Signal Processing, Ministry of Education, School of Electronics and Information Engineering, Anhui University, Hefei 230601, China.

Zhenhua Wu and Jie Chen are with the Key Laboratory of Intelligent Computing and Signal Processing, Ministry of Education, School of Electronics and Information Engineering, Anhui University, Hefei 230601, China, and also with the 38th Research Institute, China Electronics Technology Group Corporation, Hefei 230088, China.}}


\maketitle

\begin{abstract}
As remote sensing imaging technology continues to advance and evolve, processing high-resolution and diversified satellite imagery to improve segmentation accuracy and enhance interpretation efficiency emerg as a pivotal area of investigation within the realm of remote sensing. Although segmentation algorithms based on CNNs and Transformers achieve significant progress in performance, balancing segmentation accuracy and computational complexity remains challenging, limiting their wide application in practical tasks. To address this, this paper introduces state space model (SSM) and proposes a novel hybrid semantic segmentation network based on vision Mamba (CVMH-UNet). This method designs a cross-scanning visual state space block (CVSSBlock) that uses cross 2D scanning (CS2D) to fully capture global information from multiple directions, while by incorporating convolutional neural network branches to overcome the constraints of Vision Mamba (VMamba) in acquiring local information, this approach facilitates a comprehensive analysis of both global and local features. Furthermore, to address the issue of limited discriminative power and the difficulty in achieving detailed fusion with direct skip connections, a multi-frequency multi-scale feature fusion block (MFMSBlock) is designed. This module introduces multi-frequency information through 2D discrete cosine transform (2D DCT) to enhance information utilization and provides additional scale local detail information through point-wise convolution branches. Finally, it aggregates multi-scale information along the channel dimension, achieving refined feature fusion. Findings from experiments conducted on renowned datasets of remote sensing imagery demonstrate that proposed CVMH-UNet achieves superior segmentation performance while maintaining low computational complexity, outperforming surpassing current leading-edge segmentation algorithms.
\end{abstract}


\section{Introduction}
\IEEEPARstart{S}{emantic} segmentation, a key research area in interpreting remote sensing images, is extensively utilized in tasks including urban development project \cite{b1,b2,b3}, terrain categorization \cite{b4,b5,b6}, strategic land resource allocation \cite{b7}, and road network extraction \cite{b8}. With the continuous improvement in imaging resolution and the increasing width of imaging swaths, the volume and complexity of the acquired remote sensing images have significantly increased. This imposes higher demands on segmentation algorithms to more comprehensively capture target information and improve interpretation efficiency. Traditional segmentation algorithms, including Support Vector Machines (SVM) \cite{b9}, Random Forests (RF) \cite{b10}, and Conditional Random Fields (CRF) \cite{b11}, exhibit restricted capacity in target information extraction when processing high-definition remote sensing images, making it difficult to ensure segmentation accuracy. Moreover, these algorithms perform inadequately when faced with the demands of real-time image interpretation, as their computational efficiency often falls short of the demands of real-world uses.

Following the ascendancy of neural networks, deep learning methods, driven by data, automatically learn and extract more robust and discriminative features of ground objects from remote sensing images. These methods exhibit notable efficacy in the domain of remote sensing image segmentation, surpassing many traditional methods in both accuracy and efficiency. In particular, the UNet\cite{b12}, which uses a convolutional neural network (CNN) as its foundational architecture, gains widespread recognition and application within the domain of remote sensing image segmentation resulting from the exclusive U-shaped encoder-decoder structure and skip connection design. A host of further research on deep segmentation algorithms improve upon the distinctive U-shaped architecture of UNet to capture richer information and enhance segmentation accuracy. With the rise of Transformer models, their advantages in global information modeling are fully explored and applied to the field of image segmentation\cite{b13,b14,b15,b16}. Segmentation algorithms based on the Transformer framework effectively address the issue of insufficient global information extraction caused by the limited receptive field of CNNs, further improving segmentation accuracy. Swin-UNet \cite{b17}, which combines the Swin Transformer \cite{b18} with a U-shaped architecture, is the first U-shaped model purely based on Transformers. However, the self-attention mechanism in the Transformer architecture results in excessively high computational complexity, making it difficult for the computational performance of the model to meet the requirements of practical applications. To address this issue, researchers combine CNNs with Transformers to balance model accuracy and computational complexity. UNetFormer \cite{b19} leverages a combination of a CNN encoder and a Transformer decoder to adeptly capture global and local contexts. Although this approach effectively avoids the issue of excessive computational complexity, the use of CNNs in the encoder hinders the model from achieving a higher level of accuracy. The limitations of these models motivate us to develop a new framework that efficiently captures global context while preserving a linear relationship in computational complexity.

Recently, the Mamba framework \cite{b20}, which is grounded in the State Space Model (SSM), becomes a promising alternative due to its ability to capture distant dependencies while maintaining linear computational complexity. Following this, Vim \cite{b21} and VMamba \cite{b22} further extend the advantages of SSM to the discipline of image processing, providing new momentum for technological advancements in this area. Particularly within the domain of remote sensing image analysis, researchers deeply explore the application potential of SSM. For example, research on Rs3Mamba \cite{b23} and CM-UNet \cite{b24} proves that VMamba can serve as a viable substitute for conventional CNNs and Transformers in segmenting remote sensing images. These studies show that VMamba not only captures global contextual information but also maintains lower computational complexity, introducing a contemporary and efficient solution for the partitioning of remote sensing images. Although these methods take advantage of the strengths of VMamba in global information extraction to some extent, they often overlook the limitations in local information extraction that may arise due to specific scanning directions. Therefore, there is a requirement for further investigation to improve the structure of VMamba, enhancing its ability to extract local detail information while maintaining its efficiency in capturing global information, in order to formulate advanced semantic segmentation methods that are precise and efficient for remote sensing images.

In the U-shaped architecture, the method of feature fusion within the skip connections is also a critical element influencing the efficacy of U-shaped network architectures, in addition to improving efficiency and segmentation accuracy by applying VMamba in encoding-decoding architectures. Conventional skip connections merge features by linking low-level encoder features directly to high-level decoder features. However, this simple concatenation method fails to fully exploit the complementarity between features at different levels, limiting the ability of the network to differentiate between nuanced low-level information and high-level semantic aspects. As a result, the features become coarsely integrated, causing the loss of minor object details. To resolve these concerns, researchers increasingly focus on the significance of attention mechanisms and multi-scale features in semantic segmentation. The AFF module \cite{b25} and Multiattention Network (MANet) \cite{b26} demonstrate that multi-scale feature fusion strategies based on attention mechanisms effectively address the issues in skip connections. However, ECANet \cite{b27} points out that the dimensionality reduction in fully connected layers within attention mechanisms leads to information loss. Additionally, FCANet \cite{b28} shows that the traditional global average pooling method causes significant information loss. Therefore, overcoming issues in attention mechanisms and designing a feature fusion strategy that enhances the completeness and transfer precision of data is crucial. This strategy aims to achieve sophisticated merging of intricate remote sensing images and to enhance the accurate segmentation of minor objects, which represents a key challenge in current research on remote sensing semantic segmentation methods.

To tackle the previously mentioned difficulties, this paper proposes a novel hybrid semantic segmentation network based on VMamba for remote sensing image segmentation, called CVMH-UNet. The core of CVMH-UNet is the cross-scanning visual state space block (CVSSBlock). This block modifies the scanning method of Vision Mamba, changing the original SS2D, which consists of four paths—horizontal, vertical, and their reverse directions—into CS2D, which consists of four fully cross-scanning paths: horizontal, vertical, diagonal, and anti-diagonal. This allows it to capture global information more comprehensively from multiple directions. Additionally, by integrating convolutional branches, it overcomes the limitations of Vision Mamba in local information extraction, enabling in-depth exploration of both global and local features. In addition, to enhance the integrity of information and the precision of information transmission in feature fusion, this paper proposes a multi-frequency multi-scale feature fusion block (MFMSBlock). This block employs a structure with two branches to acquire information at varying scales. Within the global pathway, channel attention is treated as a compression problem, utilizing discrete cosine transform (DCT) to compress the channels. By introducing multiple frequency components, it provides global feature channel attention to enrich the representation of the channels and enhance information integrity. At the same time, it directly calculates channel weights using one-dimensional convolution with adaptive kernel sizes(Adaptive 1D Conv), avoiding the information loss that could be caused by the dimensionality reduction in fully connected layers, thus ensuring the accuracy of information transmission. In the local branch, point-wise convolution is used to provide local feature channel attention at a different scale from the global branch, capturing complex details and more extensive structural information. The two branches aggregate multi-scale information along the channel dimension, ensuring effective utilization of features at all levels. This allows both lower-level and higher-level features to be fully utilized and complementary in the fusion process, achieving refined feature fusion. The primary contributions of this research are as outlined below:

\begin{itemize}
  \item{A novel hybrid semantic segmentation network based on VMamba, called CVMH-UNet, is proposed. This network uses the CVSSBlock as the basic building unit for both the encoder and decoder, and employs a Multi-Frequency Multi-Scale Feature Fusion module in the skip connections. This achieves comprehensive extraction of both global and local features, along with refined fusion in the skip connections. Experiments on classic remote sensing datasets demonstrate that this method achieves high segmentation accuracy while keeping computational complexity low, striking a good balance between segmentation accuracy and computational efficiency.}
  \item{A new feature extraction block based on Vision Mamba, called CVSSBlock, is designed. This module comprehensively extracts global information from multiple directions using the CS2D scanning strategy and provides additional local detail information through integrated convolutional branches, ensuring adequate extraction of image information. As a consequence, the capabilities of the model are elevated while maintaining low computational complexity.}
  \item{The MFMSBlock, a newly-proposed feature fusion module, aims to replace traditional skip connections. This module introduces multiple frequency components through DCT to provide global feature channel attention, enriching channel representations and enhancing information integrity. It also uses Adaptive 1D Conv to directly compute channel weights, reducing information loss and improving the precision of information transmission. Additionally, it provides local feature channel attention at another scale through an extra point-wise convolutional branch to capture complex details and more extensive structural information. This achieves efficient aggregation of deep and shallow features obtained by the encoder and decoder.}
\end{itemize}

The continuing pages of this study are outlined in the following sequence. Section \ref{s2} introduces the related work surveyed in this study. Section \ref{s3} details the architecture of the proposed approach and the particular setup of each component. Section \ref{s4} presents the experimental setup and technical metrics. Section \ref{s5} affirms the superiority of the proposed method employing two classic datasets. Finally, Section \ref{s6} provides a summary of the research findings and insights from the study.

\section{Related Work}\label{s2}
\subsection{Vision State Space Models}
Recently, SSM become a research hotspot. As the current leading SSM, Mamba not only excels in long-range modeling capabilities but also demonstrates linear complexity in handling input sizes, addressing the computational efficiency issues of Transformers in modeling long sequences of state spaces. In the field of visual research, Vim \cite{b21} proposes a pure vision backbone model based on SSM, introducing Mamba into the visual domain for the first time. VMamba \cite{b22} enhances visual processing capabilities by introducing the Cross-Scan module, enabling the model to selectively scan images in two dimensions and demonstrating superiority in image classification tasks. LocalMamba \cite{b30} focuses on a windowed scanning strategy for visual spatial models, optimizing visual information to capture local dependencies and introducing a dynamic scanning method to search for optimal choices at different layers. In downstream visual tasks, Mamba also sees extensive use in the domain of remote sensing image segmentation research. Rs3Mamba \cite{b23} proposes using the VSSBlock to provide additional global information, aiding the convolution-centered main stream in the extraction of features, thereby enhancing global awareness of the model while preserving local details. In CM-UNet \cite{b24}, a CNN encoder is tasked with acquiring detailed image features, and a Mamba decoder is in charge of combining global information. The inclusion of an MSAA module allows for the merging of features at various scales, competently capturing long-distance correlations and multiscale global context information in high-resolution remote sensing images. MambaHSI \cite{b31} introduces a new Mamba-based model for HSI classification, MambaHSI, which simultaneously models long-distance interactions across the entire image and adaptively integrates spatial and spectral data in a flexible manner. Although these strategies utilize the strong ability of VMamba to model dependencies over long distances while maintaining linear computational complexity, they overlook the limitations in local information extraction that may arise due to specific scanning directions.

\subsection{Attention Mechanisms in Deep Learning}
SENet \cite{b32} introduces the channel attention mechanism, which performs global average pooling (GAP) along the channel dimension and uses fully connected layers to compute the weight of each channel. Due to its significant performance improvements, attention modules gain widespread attention. Subsequent studies, such as CBAM \cite{b33} and scSE \cite{b34}, use 2D convolution with a kernel size of $k\times k$ to calculate spatial attention and integrate it with channel attention. SRM \cite{b35} suggests using GAP with global standard deviation pooling. Although these attention mechanisms perform well in practice, the dimensionality reduction in the fully connected layers of GAP negatively impacts the attention mechanism. Therefore, ECANet \cite{b27} proposes using a Adaptive 1D Conv to replace the fully connected layer, avoiding the negative effects of dimensionality reduction and capturing local cross-channel interactions. Additionally, a pivotal aspect of channel attention mechanisms is the calculation of scalar values for individual channels. Due to its uncomplicated design, GAP struggles to effectively extract complex input features. FCANet \cite{b28} demonstrates that GAP represents a unique instance of the Discrete Cosine Transform (DCT) and explores different frequency component combinations within a multi-spectral framework, which enhances the ability of GAP to capture complex information. These improved attention mechanisms, by more precisely handling inter-channel and intra-channel interactions, help models better understand and utilize input data features. Therefore, effectively applying these methods in feature fusion strategies becomes key to achieving refined feature fusion.

\subsection{Skip Connections in Deep Learning}
In U-shaped networks, long skip connections are an important component. They help the network obtain detailed semantic features by bridging finer details from lower layers with coarse resolution and advanced semantic features. Although skip connections are widely used to combine features from various paths, the fusion of these connected features is typically achieved through addition or concatenation. This approach ignores the differences between the features, assigning the same weight to all of them, which prevents the complementary nature of features at different levels from being fully utilized. As a result, the capability of the network to recognize both minor details and advanced semantic features is limited. With the widespread application of attention mechanisms, SKNet \cite{b36} introduces a method for specialized feature fusion based on attention mechanisms that uses a non-linear approach. However, its constraints are apparent in its focus only on soft feature selection within the same layer, overlooking the challenge of integrating features across layers in skip connections. Moreover, the scale variation of objects represents a significant challenge within the realm of computer vision. To mitigate the issues arising from scale variation and small objects, as well as to enhance the effectiveness of skip connections, the AFF module \cite{b25} proposes a unified and generalized method for feature fusion using attention mechanisms. Similarly, the Multiattention Network (MANet) \cite{b26} extends attention mechanisms across multiple scales, aggregating contextual information at different scales to obtain a more complete and nuanced merging of features. Although these feature fusion methods can effectively enhance the feature fusion effects of long skip connections, they often overlook the issues of information transmission loss and insufficient information utilization in the attention mechanisms, as mentioned in ECANet \cite{b27} and FCANet \cite{b28}.

\section{Methodology}\label{s3}

\begin{figure*}[t]
  \centering
  \includegraphics[width=6in]{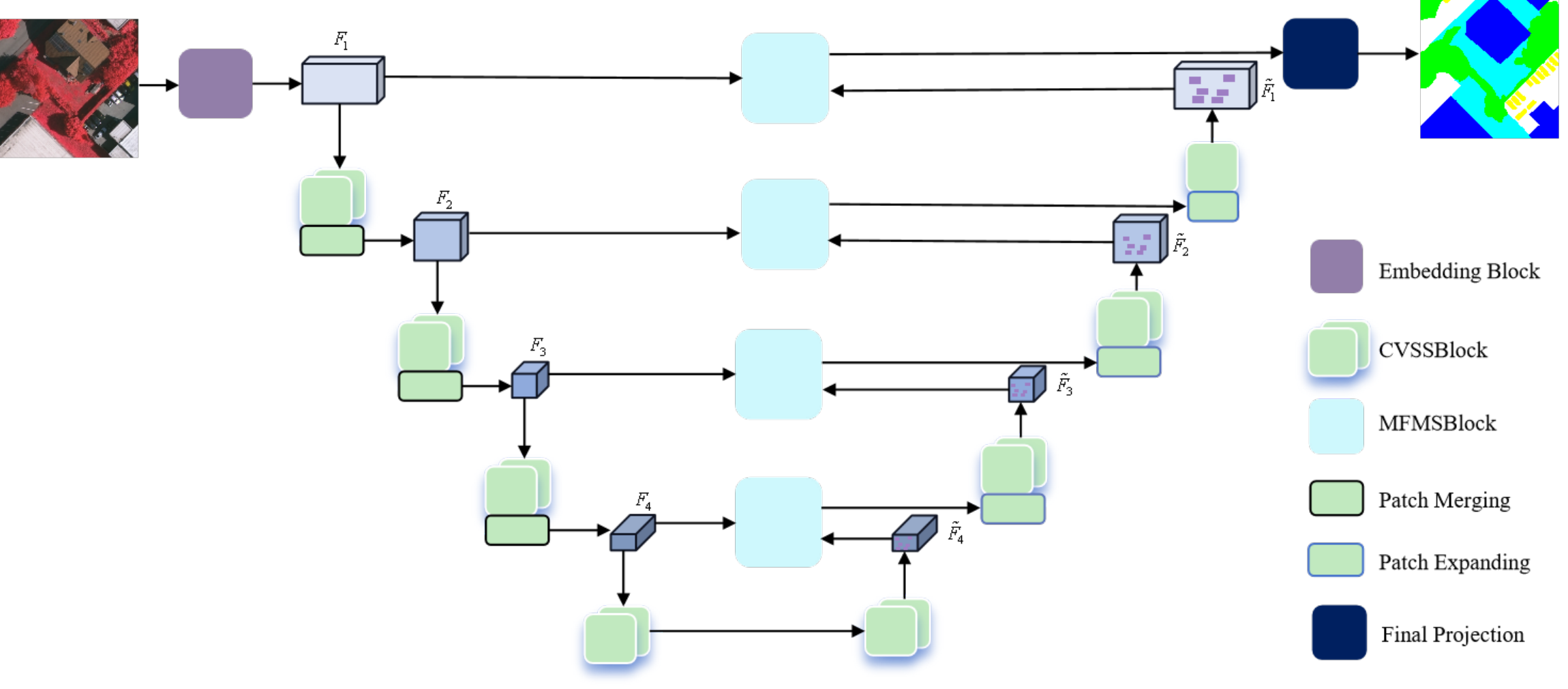}
  \caption{The overall architecture of CHVM-UNet.}
  \label{fig1}
\end{figure*}

\subsection{Overall Architecture}
The complete network structure of the proposed CVMH-UNet is depicted in Fig.~\ref{fig1}, which is designed and improved based on VMamba. The CVMH-UNet architecture primarily comprises an embedding block, an encoder, a decoder, a multi-frequency multi-scale feature fusion block (MFMSBlock), and a final projection layer. The original remote sensing image $x\in\mathbb{R}^{3\times H\times W}$ is segmented into non-intersecting $4\times4$ patches in the embedding block and mapped to a C-dimensional feature space. Subsequently, it undergoes processing by a normalization layer to produce the embedded feature map $x'\in\mathbb{R}^{C\times\frac H4\times\frac W4}$, where the mapping dimension C is set to 96. Subsequently, the embedded features are fed into the encoder, which consists of CVSSBlocks, to extract features. In the first three stages, patch merging is employed for downscaling to progressively decrease spatial extents and acquire hierarchical features. Afterward, the decoder, also composed of CVSSBlocks, reconstructs the features, and in the last three stages, patch expanding is used for upsampling to increase spatial dimensions and further extract higher-level feature representations. Finally, the final result is obtained through the final projection layer, restoring the dimensions of the output image to match those of the input image. It is important to note that both the encoder and decoder are composed of CVSSBlocks, but they follow an asymmetric configuration strategy. In the four stages of the encoding part, the number of CVSSBlocks is [2,2,2,2], while in the four stages of the decoding part, the number of CVSSBlocks is [2,2,2,1]. This design aims to balance model complexity and computational efficiency. Additionally, to better restore details and capture more target information, the decoder interfaces with the encoder through the MFMSBlock for feature fusion. This ensures strong information integrity and precision in information transmission, strengthening the discriminative capacity of the features and refining detailed fusion of small objects, leading to segmentation results that are more accurate and exact.

\subsection{CVSSBlock}
\begin{figure*}[t]
  \centering
  \includegraphics[width=5in]{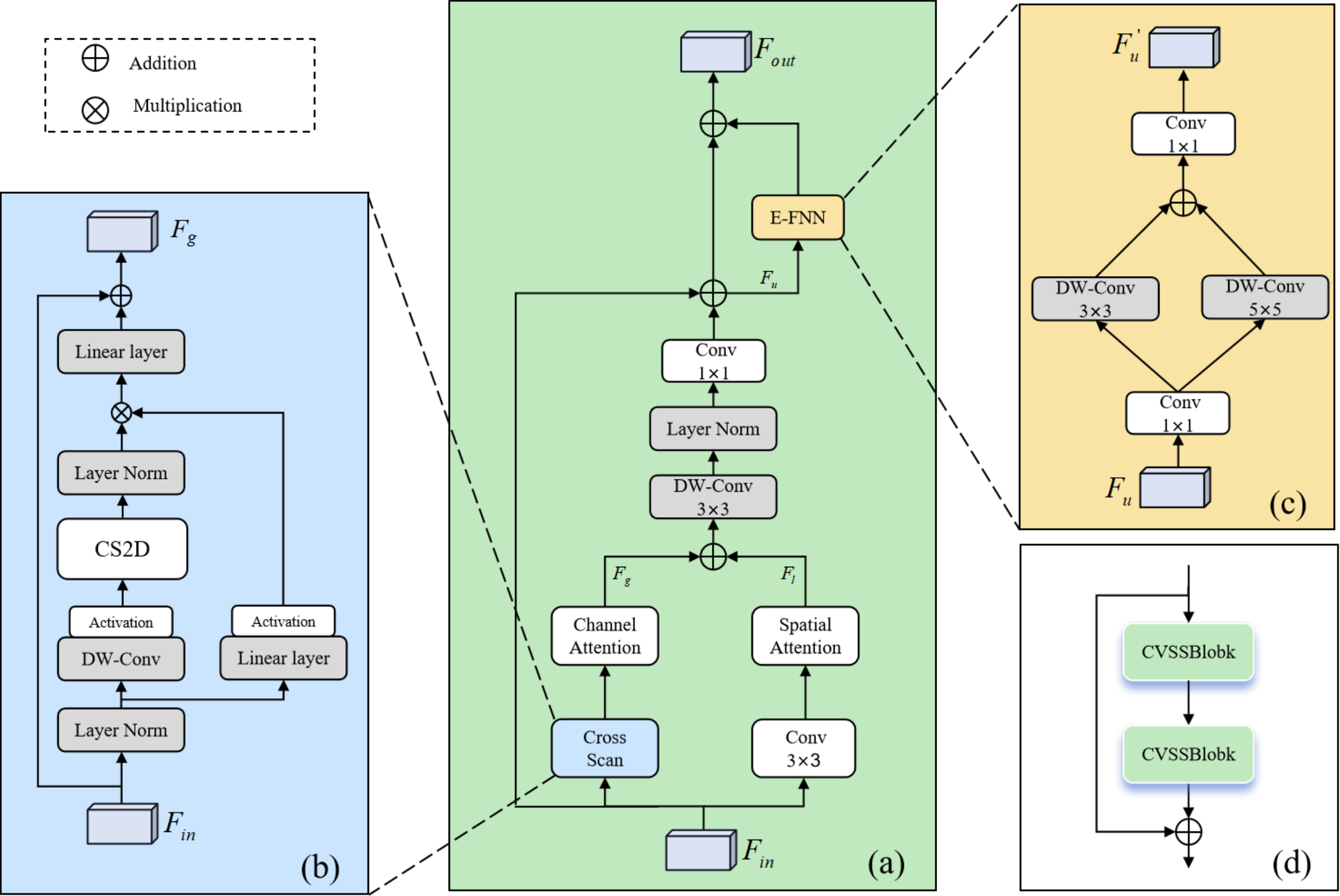}
  \caption{(a) Overall architecture of the CVSSBlock, (b) Detailed structure of the Cross Scan, (c) Detailed structure of the E-FNN, (d) Connection method of the CVSSBlock.}
  \label{fig2}
\end{figure*}

The basic unit of the CVMH-UNet encoder-decoder architecture is the proposed CVSSBlock, which is an improved version of the VSSBlock from VMamba. The advantage of this module lies in its higher feature representation capability, achieved through the integration of convolutional branches and optimized scanning methods. The specific configuration is illustrated in Fig.~\ref{fig2}(a). Inspired by VMamba \cite{b22}, and offset the deficiency in local information extraction resulting from the SS2D scanning method in the VSSBlock, we designed a structure with two branches in the CVSSBlock by adding a convolutional branch to capture local information. Additionally, to address the inefficiency in information extraction caused by the SS2D method in VMamba, which scans in different directions along the same path, we designed the CS2D method with intersecting scanning paths. This allows the extraction of spatial features from multiple directions without increasing computational complexity, making it more efficient in capturing global information from high-resolution remote sensing images.
\begin{figure*}[!t]
  \centering
  \includegraphics[width=5in]{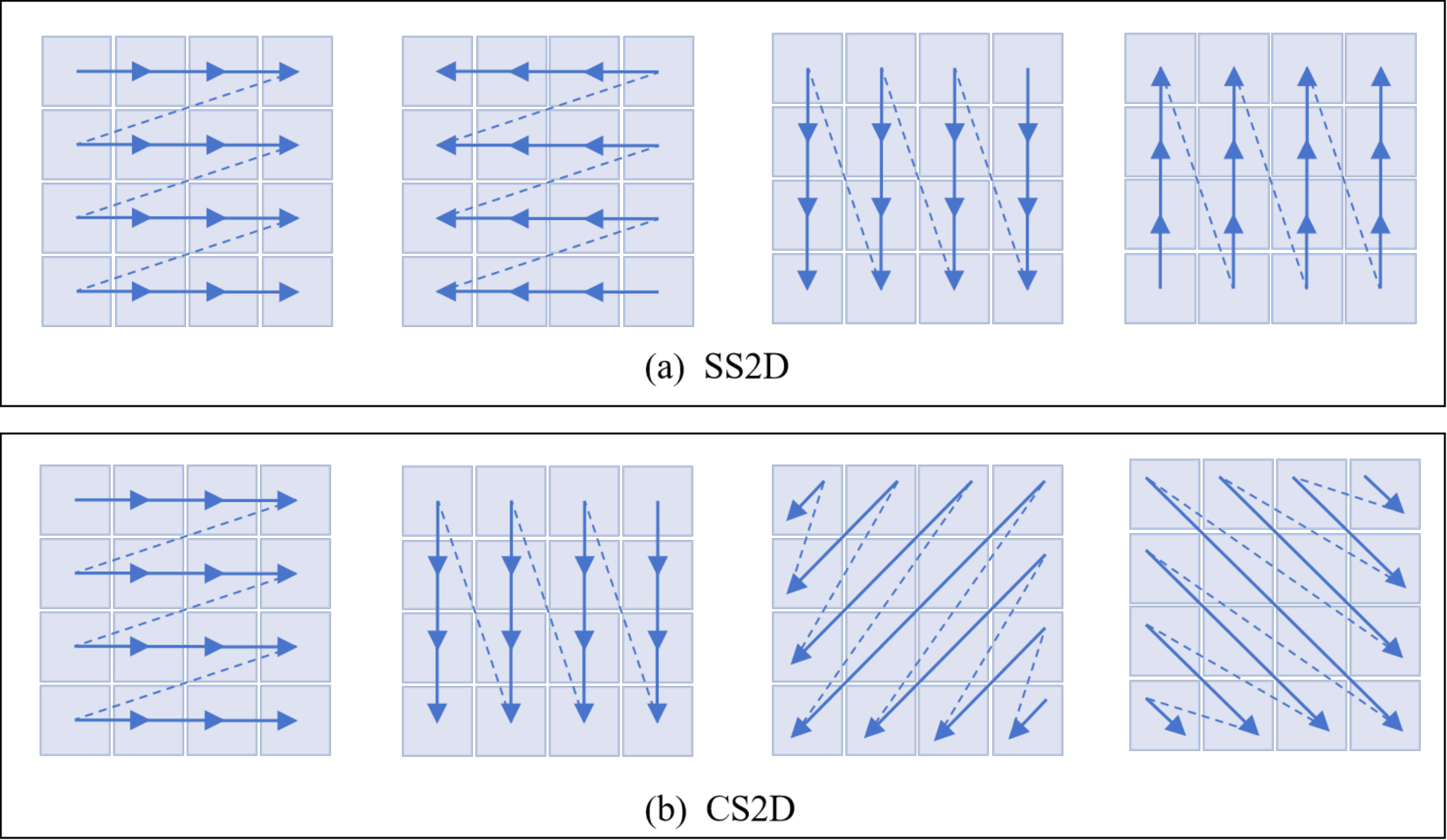}
  \caption{Comparison of two different scanning methods, (a) Four scanning paths of SS2D, (b) Four scanning paths of CS2D.}
  \label{fig3}
\end{figure*}

As can be seen in Fig.~\ref{fig2}(a), two branches are applied on the input feature $F_{in}$ to obtain the global and local information. More specifically, within the global pathway, the input features first go through the Cross Scan module (CS) to extract the global information, and then the global information is input to the channel attention (CA) mechanism to obtain the global feature $F_{g}$. This system leverages the interdependencies among channel mappings to enhance the representation of specific semantic features. In the CS module, after layer normalization, input $F_{in}$ is channeled into two distinct branches. The initial branch includes a linear layer and an activation step. The other branch encompasses a linear layer, a depthwise separable convolution, and an activation step, before entering the CS2D module for extracting global features across various directions. Post this, the features undergo normalization and are subsequently merged element-wise with the output from the global pathway. In conclusion, a linear layer blends the features, which are subsequently integrated with the residual connection to produce the final output of the CS module. In the local branch, the input features first pass through a $3\times3$ convolution to extract local information, which is then enhanced by the Spatial Attention (SA) mechanism to emphasize local details and suppress irrelevant regions, resulting in the local feature $F_{l}$. To sum up, the process can be represented as follows:
\begin{figure*}[t]
  \centering
  \includegraphics[width=5in]{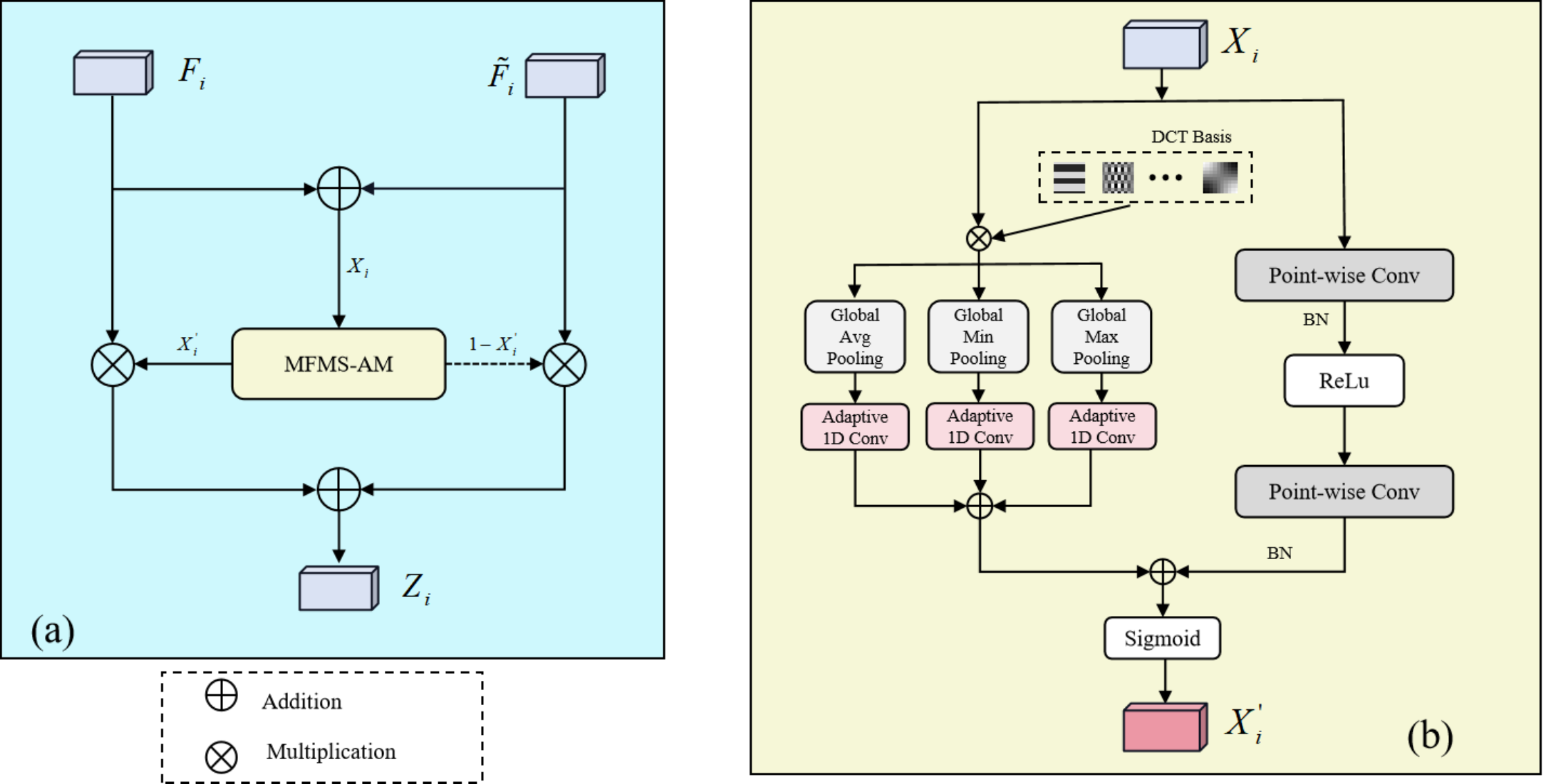}
  \caption{(a) Structure of MFMSBlock (b) Structure of MFMS-AM.}
  \label{fig4}
\end{figure*}

\begin{equation}
\label{eq5}
  F_g=CA(CS(F_{in}))
\end{equation}
\begin{equation}
  \label{eq6}
  F_l=SA(Conv_{3\times3}(F_{in}))
\end{equation}
where $Conv_{3\times3}(\cdot)$ represents a $3\times3$ convolution, $CS(\cdot)$ represents the Cross Scan module, and $CA(\cdot)$ and $SA(\cdot)$ represent the channel attention mechanism and spatial attention mechanism, respectively. Afterward, the global and local features are added together, followed by depthwise separable convolution, layer normalization, and a $1\times1$ convolution to refine the combined features. These refined features are then are merged with the starting input $F_{in}$ through a residual connection, generating the fused feature $F_{u}$. Finally, to enhance feature representation, the E-FFN module is introduced to further improve information interaction between different channels, achieving better segmentation performance. This process can be represented as follows:
\begin{equation}
\label{eq7}
  F_{u}=F_{in}+Conv_{1\times1}(LN(dwConv_{3\times3}(F_{g}+F_{l})))
\end{equation}
\begin{equation}
\label{eq8}
  F_{out}=F_u+EN(F_u)
\end{equation}
where $F_{out}$ denotes the output features of the CVSSBlock, $dwConv_{3\times3}(\cdot)$ denotes the $3\times3$ depth-separable convolution, $LN(\cdot)$ denotes layer normalization, $Conv_{\mathrm{l\times l}}(\cdot)$ denotes the $1\times1$ convolution, and $EN(\cdot)$ denotes the E-FNN module.

Additionally, to optimize the feature extraction process and reduce information loss, enabling the network to better extract features at each layer, two CVSSBlocks are connected using a residual connection. As shown in Fig~\ref{fig2}(d), this approach reduces information loss, helping the network better identify and segment targets in high-resolution remote sensing images. Moreover, it assists the network in learning more refined feature representations, thus enhancing the precision of the segmentation of the model.

\subsection{MFMSBlock}
The encoder and decoder of CVMH-UNet are capable of extracting low-level features with rich texture details and high-level features with semantic information, respectively. To more effectively utilize this information, we designed the MFMSBlock. The core idea is to introduce frequency information into global pooling and combine it with additional local scale analysis to construct a multi-frequency, multi-scale feature fusion mechanism. This module adopts a structure with two branches to acquire information across different scales different scales. Within the global branch, 2D DCT is used to introduce frequency information into global pooling, and various global pooling techniques are utilized to capture fine features across different frequency bands, enhancing the completeness of the information. Additionally, Adaptive 1D Conv \cite{b27} is employed to directly compute channel attention, preventing data loss due to the dimensional reduction by fully connected layers, thus improving the precision of information transmission and producing more accurate global feature channel attention. In the local branch, point-wise convolution is used to provide local feature channel attention at another scale, capturing complex details and broader structural information. Finally, the two branches aggregate information from different scales along the channel dimension. The fusion approach not only strengthens the capacity of the model to detect local texture details but also enhances its comprehension of global semantic frameworks, substantially elevating the overall efficacy of the model.

The overall structure of the proposed MFMSBlock is depicted in Fig.~\ref{fig4}(a). The feature map $X_{i}$, obtained by adding the low-level feature map $F_{i}$ from the ${i}$-th layer of the encoder and the high-level semantic feature map $\tilde{F}_i$ from the ${i}$-th layer of the decoder, passes through the multi-frequency multi-scale attention mechanism (MFMS-AM) to generate fusion weights $X_{i}^{\prime}$ and $1-X_{i}^{'}$. This allows the MFMSBlock to conduct a soft selection process or a weighted mean between $F_{i}$ and $\tilde{F}_i$. This module can be expressed as:
\begin{equation}
\label{eq9}
  Z_i=MA(F_i+\tilde{F}_i)\times F_i+(1-MA(F_i+\tilde{F}_i))\times\tilde{F}_i
\end{equation}
where $Z_{i}\in\mathbb{R}^{C\times H\times W}$ represents the fused features, and $MA(\cdot)$ represents the attention weight values generated by MFMS-AM.

The overall structure of MFMS-AM is shown in Fig.~\ref{fig4}(b). It uses 2D DCT to express image features as a weighted sum of cosine functions at different frequencies. The representation at the ${k}$-th frequency is denoted as $X_i^k$, and is calculated using the following formula:
\begin{equation}
\label{eq10}
  X_i^k=\sum_{h=0}^{H-1}\sum_{w=0}^{W-1}\left(X_i\right)_{:,h,w}D_{h,w}^{u_k,v_k}
\end{equation}
where $(u_{k},v_{k})$ is the index of the frequency corresponding to $X_i^k$. In addition, the 2D DCT basis image using the top-K selection strategy \cite{b28} is defined as:
\begin{equation}
\label{eq11}
  D_{h,w}^{u_k,v_k}=\cos\Bigl(\frac{\pi h}{H}\Bigl(u_k+\frac{1}{2}\Bigr))\cos\bigl(\frac{\pi w}{W}\bigl(\nu_k+\frac{1}{2}\bigr)\bigr)
\end{equation}

Next, $X_i^k$ is compressed by applying global average pooling, global max pooling, and global min pooling, respectively. These are then passed through a Adaptive 1D Conv \cite{b27} to avoid dimensionality reduction and directly compute the weights. With the channel dimension $C$ as a reference, the kernel size ${\phi}$ can be dynamically calculated by applying the following equation:
\begin{equation}
\label{eq12}
  \phi=\left|\frac{\log_2(C)}{\alpha}+\frac{\beta}{\alpha}\right|_{odd}
\end{equation}
where $\begin{vmatrix}{\lambda}\end{vmatrix}_{odd}$ denotes the nearest odd number to ${\lambda}$. In the experiments conducted for this paper, ${\alpha}$ and ${\beta}$ are uniformly set to 2 and 1, respectively. Finally, the branches are summed to obtain the global channel attention map $G(X_{i})\in\mathbb{R}^{C}$. Additionally, the proposed MFMS-AM follows the concept of the Multi-Scale Channel Attention Module (MS-CAM) \cite{b25}, where an additional point-wise convolution branch provides local feature channel attention at another scale. It only leverages point-wise interactions at each spatial location to enable channel interaction. The local feature channel attention map $L(X_{i})\in\mathbb{R}^{C\times H\times W}$ is acquired via the subsequent bottleneck architecture:
\begin{equation}
\label{eq13}
  L(X_i)=\psi\Big(pwCon\nu_2\Big(\theta\Big(\psi\Big(pwCon\nu_1(X_i)\Big)\Big)\Big)\Big)
\end{equation}
where $pwConv_{_1}(\cdot)$ and $pwConv_{_2}(\cdot)$ represent point-wise convolutions, $\theta(\cdot)$ denotes the rectified linear unit (ReLU), and $\psi(\cdot)$ denotes batch normalization (BN). It is worth noting that $L(X_{i})$, which aligns with the shape of the input features, allows it to preserve and highlight subtle details in low-level features. Given $G(X_{i})$ and $L(X_{i})$, MFMS-AM can produce the fusion weight $X_i'$ through the activation function (Sigmoid):
\begin{equation}
\label{eq14}
  X_i'=Sigmoid(G(X_i)+L(X_i))
\end{equation}

The introduction of multi-frequency information improves the integrity of the information, and the Adaptive 1D Conv \cite{b27} enhances the precision of information transmission. This allows the MFMSBlock to more effectively utilize the feature information produced by the encoder and decoder. In addition, the point-wise convolution branch provides extra scale information, enabling the MFMSBlock to better handle targets of different scales in high-resolution remote sensing images. Applying these strategies ensures that both low-level and high-level features are fully utilized and complement each other during the fusion process, resulting in refined feature fusion.

\subsection{Remote Sensing Image Segmentation Based on CVMH-UNet}
Preparing the training set required for network training, including the original remote sensing images and their corresponding labels. Next, preparing the data by adjusting the image dimensions to fulfill the network's input criteria, applying data augmentation and normalization, and generating pixel-level labels indicating the class to which each pixel belongs. Once the data processing is complete, input the processed data into CVMH-UNet. First, feature extraction is performed through the encoder composed of CVSSBlocks, followed by feature reconstruction through the decoder, which is also composed of CVSSBlocks. The MFMSBlock is then used to fuse the features generated by each layer of the encoder and decoder. Finally, in the final projection layer, the output channel count is modified to correspond with the class count, predicting the probability of each pixel belonging to a given class.

Throughout the training process, the performance of the predicted feature maps is judged against the ground truth labels using multi-class cross-entropy and Dice loss as metrics. Backpropagation is then performed to adjust the model parameters, ensuring the model conforms more precisely to the training data. After setting a reasonable number of iterations, the weights of the model are preserved once the loss function stabilizes, and these weights are used for predictions.

\section{Dataset and Experimental Setting}\label{s4}
\subsection{Datasets}
\subsubsection{ISPRS Vaihingen}The dataset consists of 16 high-resolution remote sensing images with a resolution of 9 cm, each image averaging approximately $2500\times2000$ pixels and composed of RGB images with green, red, and near-infrared channels. The dataset includes five foreground classes: impervious surfaces, buildings, low vegetation, trees, and cars, along with one background class (clutter). The training set is constituted by images with index numbers 1, 3, 23, 26, 7, 11, 13, 28, 17, 32, 34, and 37 out of the 16 images. The test set consists of images with index numbers 5, 21, 15, and 30.
\subsubsection{ISPRS Potsdam}The dataset consists of 16 high-resolution remote sensing images with a resolution of 9 cm, each image averaging approximately $6000\times6000$ pixels and composed of RGB images with blue, red, and green channels. In alignment with the Vaihingen dataset, it includes six classes: impervious surfaces, buildings, low vegetation, trees, cars, and one background class (clutter). We use the images with IDs: $2_{-}13$, $2_{-}14$, $3_{-}13$, $3_{-}14$, $4_{-}13$, $4_{-}14$, $4_{-}15$, $5_{-}13$, $5_{-}14$, $5_{-}15$, $6_{-}13$, $6_{-}14$, $6_{-}15$, and $7_{-}13$ as the test set, while The training set is constituted by 23 images, not including image $7_{-}10$, as it contains errors in its annotations.

\setkeys{Gin}{width=1.0\textwidth}
\begin{table*}[t]
\belowrulesep=0pt
\aboverulesep=0pt   
\setlength{\tabcolsep}{12pt}
\caption{\textbf{Quantitative comparison on the ISPRS Vaihingen dataset. The accuracy for each class is presented in the form of IoU (\%).}}
\centering
\scalebox{1.0}{
\begin{tabular}{c|ccccc|ccc}
\toprule
Method &Imp.surf. &Building &Lowveg. &Tree &Car &mF1(\%) &mIoU(\%) & OA(\%)\\
\midrule
BANet\cite{b37} &71.57 &80.84 &52.02 &70.66 &45.54 &77.33 &64.13 &77.47\\
ABCNet\cite{b38} &77.52 &88.09 &56.94 &75.23 &61.46 &83.11 &71.85 &84.80\\
UNetFormer\cite{b19} &76.75 &87.18 &57.20 &74.49 &57.54 &82.33 &70.63 &83.74\\
${\rm A^{2}}$-FPN\cite{b39} &78.56 &87.12 &59.12 &75.12 &62.81 &83.67 &72.55 &84.41\\
MAResU-Net\cite{b40} &80.27 &\underline{89.33} &59.87 &75.50 &65.28 &84.67 &74.05 &85.19\\
MANet\cite{b26} &79.28 &88.10 &59.43 &\underline{76.14} &67.62 &84.76 &74.11 &\underline{85.37}\\
CMTFNet\cite{b41} &\underline{81.17} &89.31 &\underline{61.00} &75.86 &\underline{67.91} &\underline{85.38} &\underline{75.05} &85.04\\
Rs3Mamba\cite{b23} &79.78 &88.18 &58.70 &75.65 &63.58 &84.06 &73.18 &84.30\\
CM-UNet\cite{b24} &79.77 &89.10 &59.43 &75.62 &66.56 &84.72 &74.10 &84.55\\
CHVM-UNet(ours) &\textbf{81.92} &\textbf{90.25} &\textbf{62.11} &\textbf{77.13} &\textbf{68.44} &\textbf{85.98} &\textbf{75.97} &\textbf{85.82}\\
\bottomrule
\end{tabular}}
\label{tab1}
\end{table*}
\setkeys{Gin}{width=1.0\textwidth}
\begin{table*}[t]
\belowrulesep=0pt
\aboverulesep=0pt   
\setlength{\tabcolsep}{12pt}
\caption{\textbf{Quantitative comparison on the ISPRS Potsdam dataset. The accuracy for each class is presented in the form of IoU (\%).}}
\centering
\scalebox{1.0}{
\begin{tabular}{c|ccccc|ccc}
\toprule
Method &Imp.surf. &Building &Lowveg. &Tree &Car &mF1(\%) &mIoU(\%) & OA(\%)\\
\midrule
BANet\cite{b37} &73.80 &80.25 &63.48 &58.86 &74.23 &82.19 &70.12 &82.56\\
ABCNet\cite{b38} &81.94 &90.11 &\underline{71.89} &73.64 &82.75 &88.78 &80.07 &88.61\\
UNetFormer\cite{b19} &82.00 &89.41 &71.08 &71.18 &82.41 &88.23 &79.22 &88.16\\
${\rm A^{2}}$-FPN\cite{b39} &82.54 &90.55 &71.78 &72.76 &82.82 &88.77 &80.09 &88.69\\
MAResU-Net\cite{b40} &82.07 &\underline{90.73} &71.76 &72.36 &\underline{83.87} &88.81 &80.16 &88.75\\
MANet\cite{b26} &\underline{82.36} &\textbf{90.95} &71.66 &72.59 &83.34 &88.85 &\underline{80.23} &\textbf{89.03}\\
CMTFNet\cite{b41} &82.49 &90.48 &71.81 &\underline{73.23} &83.07 &\underline{88.86} &80.22 &88.65\\
Rs3Mamba\cite{b23} &82.17 &89.83 &71.28 &72.49 &82.76 &88.54 &79.71 &87.99\\
CM-UNet\cite{b24} &82.37 &90.66 &71.45 &72.94 &83.19 &88.79 &80.12 &88.48\\
CHVM-UNet(ours) &\textbf{83.40} &90.70 &\textbf{72.91} &\textbf{73.97} &\textbf{83.99} &\textbf{89.35} &\textbf{80.99} &\underline{88.96}\\
\bottomrule
\end{tabular}}
\label{tab2}
\end{table*}

\subsection{Experimental Setting}
The experiments are carried out on a standalone NVIDIA GeForce RTX 4080 GPU with 16GB of RAM, and using the PyTorch platform. AdamW is used as the optimizer for training all models. The training parameters include a learning rate of 0.001, a weight decay coefficient of 0.05, a batch size of 5, and a total of 300 epochs. Additionally, to ensure rigor and fairness in the experiments, all networks are tested under the same framework, using identical data processing methods, the same optimizer, and the same loss function.

The performance of the proposed CHVM-UNet is juxtaposed with that of several state-of-the-art semantic segmentation methods. The comparison involves models that employ CNN and Transformer methodologies, such as BANet \cite{b37}, ABCNet \cite{b38}, UNetFormer \cite{b19}, A2-FPN \cite{b39}, MAResU-Net \cite{b40}, MANet \cite{b26}, CMTFNet \cite{b41}, as well as Rs3Mamba \cite{b23} and CM-UNet \cite{b24}, which are based on Vision Mamba. Unless otherwise specified, the optimal results are presented in bold within the tables, and the near-optimal results are signified with underlining.

\subsection{Evaluation Metrics}
Assessing the performance of our model involves three metrics—overall accuracy (OA), mean intersection over union (mIoU), and mean F1 score (mF1)—to compare it with the most recent state-of-the-art methods. The following explanation pertains to the calculation of OA, mIoU, and mF1 using the collective confusion matrix:
\begin{gather}
  \mathrm{OA}=\frac{\sum_{k=1}^N\mathrm{TP}_k}{\sum_{k=1}^N\mathrm{TP}_k+\mathrm{FP}_k+\mathrm{TN}_k+\mathrm{FN}_k} \label{15} \\
  \mathrm{mIoU}=\frac{1}{N}\sum_{k=1}^{N}\frac{\text{TP}_{k}}{\text{TP}_{k}+\text{FP}_{k}+\text{FN}_{k}} \label{16} \\
  \mathrm{precision}=\frac{1}{N}\sum_{k=1}^{N}\frac{\mathrm{TP}_{k}}{\mathrm{TP}_{k}+\mathrm{FP}_{k}} \label{17} \\
  \mathrm{recall}=\frac{1}{N}\sum_{k=1}^{N}\frac{\mathrm{TP}_{k}}{\mathrm{TP}_{k}+\mathrm{FN}_{k}} \label{18} \\
  \mathrm{Fl}=2\times\frac{\mathrm{precision}\times\mathrm{recall}}{\mathrm{precision}+\mathrm{recall}} \label{19}
\end{gather}
where $\mathrm{TP}_{k}$, $\mathrm{FP}_{k}$, $\mathrm{TN}_{k}$, $\mathrm{FN}_{k}$ represent the true positives, false positives, true negatives, and false negatives for class k, respectively. OA represents the proportion of accurately predicted pixels out of the overall pixel count. In addition, we evaluate the model's complexity by considering its floating-point operations per second (FLOPs) and the total number of parameters. Higher values of OA, mIoU, and mF1 indicate better model performance, while lower values of Flops and Params indicate a more lightweight model.

\section{Experimental Results and Analysis}\label{s5}
\begin{figure*}[t]
  \centering
  \includegraphics[width=6in]{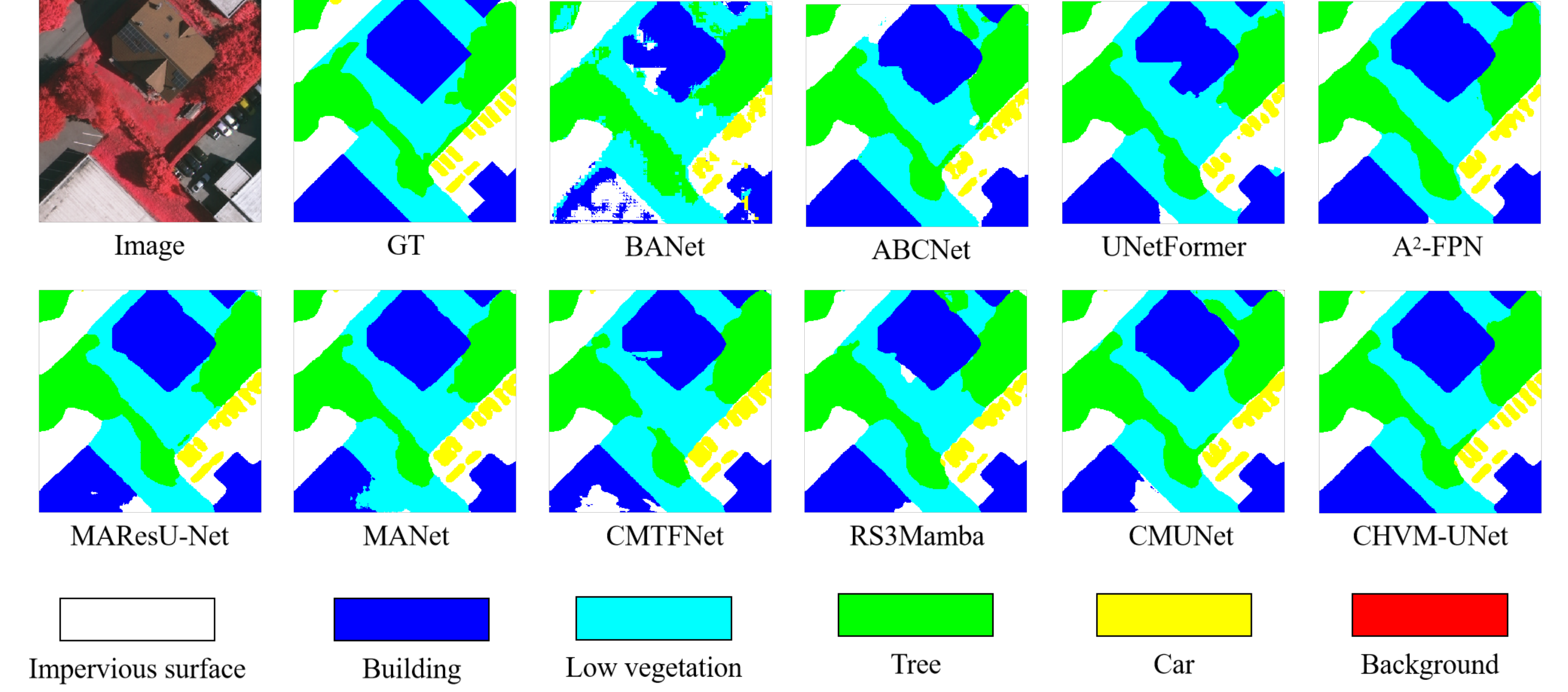}
  \caption{Segmentation results of different methods on the ISPRS Vaihingen dataset.}
  \label{fig5}
\end{figure*}
\begin{figure*}[t]
  \centering
  \includegraphics[width=6in]{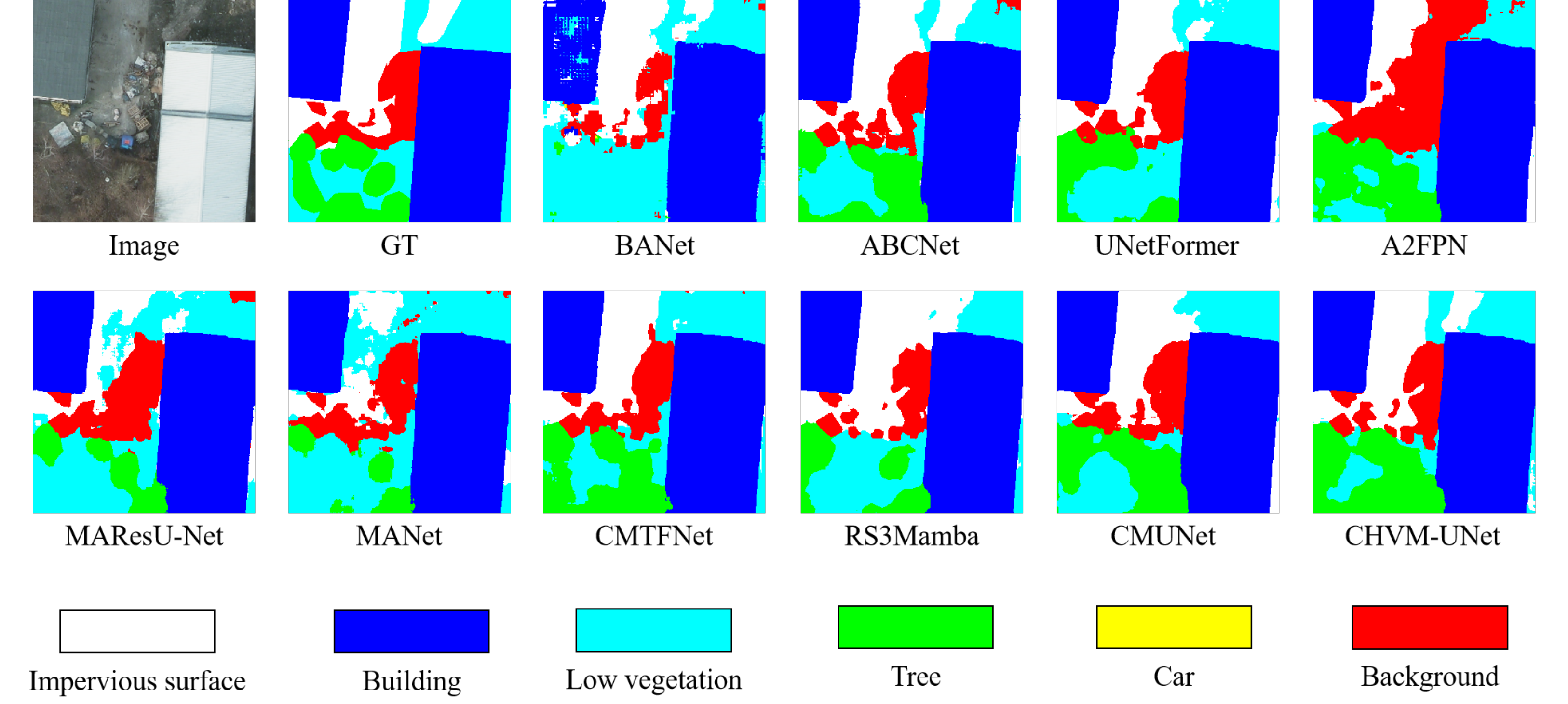}
  \caption{Segmentation results of different methods on the ISPRS Potsdam dataset.}
  \label{fig6}
\end{figure*}
\subsection{Comparison With State-of-the-Art Methods on ISPRS Vaihingen}
The comparative analysis of different methods on the ISPRS Vaihingen dataset is summarized in Table~\ref{tab1}. As the background class (clutter) represents a minor fraction of the entire dataset, consistent with other studies, we exclude it from our accuracy reporting. As shown in Table~\ref{tab1}, our proposed CVMH-UNet achieves optimal scores in the metrics of mF1, mIoU, and OA. In contrast to the next best method, mF1 improves by 0.6\%, mIoU by 0.92\%, and OA by 0.45\%. Our approach garners the top scores in every category. Notably, it achieved a 90.25\% mIoU score for the large-scale building class, surpassing the next best method by 0.92\%. The small-scale car category saw an IoU score of 68.44\%, which is 0.53\% higher than the method in second place. These results clearly demonstrate that CVMH-UNet effectively segments targets of different scales, meeting the current demands for high-resolution remote sensing image segmentation. To visually illustrate the differences between our method and others, Fig.~\ref{fig5} illustrates the segmentation outcomes from different methods on the ISPRS Vaihingen dataset. Fig.~\ref{fig5} presents the information that CVMH-UNet better preserves the complete contour of large-scale targets (building category) and more accurately identifies and segments each individual vehicle in small-scale targets (car category). In summary, CVMH-UNet demonstrates excellent segmentation capabilities for targets of varying scales, outperforming other compared methods.

\subsection{Comparison With State-of-the-Art Methods on ISPRS Potsdam}
To substantiate the effectiveness of our model, we conducted comparative experiments with other advanced methods on the larger ISPRS Potsdam dataset, with the with results tabulated in Table~\ref{tab2}. Consistent with the ISPRS Vaihingen dataset, the background class accuracy remains undisclosed. The results in Table~\ref{tab2} indicate that our method achieves the highest IoU scores in all categories except for the building category, while also obtaining the highest mF1 and mIoU scores. In contrast to the next best method, mF1 improves by 0.49\%, and mIoU increases by 0.76\%. Furthermore, to visually illustrate the differences between our method and others, a depiction of the segmentation results from various methods on the ISPRS Potsdam dataset can be seen in Fig.~\ref{fig6}. By comparing the segmentation results in Fig.~\ref{fig6}, it is evident that the proposed CVMH-UNet captures more detailed information, with segmentation results that are closer to the actual ground truth. The segmentation boundaries are more distinct, with a reduced number of misclassified pixels, thus outperforming alternative methods.

\subsection{Ablation Experiments}
\begin{table}[H]
\belowrulesep=2pt
\aboverulesep=2pt   
\setlength{\tabcolsep}{5pt}
\caption{\textbf{Results of the ablation study comparison.}}
\centering
\begin{tabular}{cc|ccc}
\hline
CVSSBlock &MFMSBlock &mIoU(\%) &FLOPs(Gbps) &Params(Mb)\\
\hline
$\times$ &$\times$ &75.08 &4.09 &22.04\\
\checkmark &$\times$ &75.59 &5.61 &30.44\\
$\times$ &\checkmark &75.42 &4.18 &22.43\\
\checkmark &\checkmark &75.97 &5.71 &30.84\\
\hline
\end{tabular}
\label{tab3}
\end{table}

\subsubsection{Effect of Each Module of CVMH-UNet}To determine the performance of individual modules within the CVMH-UNet, we carry out ablation experiments using the ISPRS Vaihingen dataset. We assessed the effectiveness of the proposed CVSSBlock and MFMSBlock by removing and adding the respective modules. Table III summarizes the results of the ablation comparisons, where "\checkmark" and "$\times$" indicate the presence and absence of the corresponding modules, respectively. For CVSSBlock, we specifically examined the effectiveness of the residual structure and the local branch within the module. For MFMSBlock, we focused on the effectiveness of the introduced multi-frequency information and the Adaptive 1D Conv. The baseline model adopts a U-Net architecture with VSSBlock as the backbone, built on the Vision Mamba framework.
\paragraph{Baseline+CVSSBlock}The CVSSBlock, constructed with integrated convolutional branches and an optimized scanning method, achieves higher feature representation capability. Table~\ref{tab3} illustrates the deployment of CVSSBlock on the ISPRS Vaihingen dataset resulted in a 0.51\% improvement in mIoU, while only slightly increasing the model's computational complexity. This indicates that our designed module can significantly enhance the feature extraction ability of the model while maintaining low computational complexity, thereby demonstrating the module's effectiveness.
\paragraph{Baseline+MFMSBlock}The MFMSBlock achieves effective skip connection fusion by constructing a multi-frequency, multi-scale feature fusion mechanism. As shown in Table~\ref{tab3}, MFMSBlock leads to a 0.34\% improvement in mIoU on the ISPRS Vaihingen dataset at a very low computational cost, which solidly establishes the module's lightweight architecture and its proven effectiveness.
\paragraph{Baseline+CVSSBlock+MFMSBlock (CVMH-UNet)}Combining CVSSBlock and MFMSBlock constructs the complete CVMH-UNet. As shown in Table~\ref{tab3}, the combination of the two modules yields the highest mIoU score. Compared to the unmodified Baseline, CVMH-UNet significantly improves segmentation accuracy at the cost of only a slight increase in computational complexity. This demonstrates that the proposed CVSSBlock and MFMSBlock not only fulfil the needs of high-resolution remote sensing image segmentation but also satisfy the demands of practical applications.

\subsubsection{Effect of CVSSBlock}
\begin{table}[H]
\belowrulesep=2pt
\aboverulesep=2pt   
\setlength{\tabcolsep}{6pt}
\caption{\textbf{Comparison of different scanning strategies.}}
\centering
\begin{tabular}{cc|ccc}
\hline
SS2D &CS2D &mIoU(\%) &FLOPs(Gbps) &Params(Mb)\\
\hline
\checkmark &$\times$ &75.08 &4.09 &22.04\\
$\times$ &\checkmark &75.30 &4.09 &22.04\\
\hline
\end{tabular}
\label{tab4}
\end{table}
\begin{table}[H]
\belowrulesep=2pt
\aboverulesep=2pt   
\setlength{\tabcolsep}{4pt}
\caption{\textbf{Ablation study results for each module of the CVSSBlock.}}
\centering
\begin{tabular}{cc|ccc}
\hline
Local Branch &Residual Block &mIoU(\%) &FLOPs(Gbps) &Params(Mb)\\
\hline
$\times$ &$\times$ &75.30 &4.09 &22.04\\
\checkmark &$\times$ &75.42 &5.61 &30.44\\
$\times$ &\checkmark &75.44 &4.09 &22.04\\
\checkmark &\checkmark &75.59 &5.61 &30.44\\
\hline
\end{tabular}
\label{tab5}
\end{table}
To substantiate the capabilities of the novel CS2D scanning method in effectively capturing global features, to assess the effectiveness of the integrated convolutional branch in improving VMamba's ability to extract local features, and to examine the advantages of the residual structure in optimizing the feature extraction process and reducing information loss, we conducted an ablation study on these three key components. First, we assessed the comparative performance of the original SS2D scanning method versus our innovative CS2D approach. As shown in Table~\ref{tab4}, the CS2D scanning method increases the mIoU score by 0.22\% without increasing the model's computational complexity, fully demonstrating CS2D's superior ability to extract global features. Additionally, the other two main designs of CVSSBlock are the integration of the convolutional branch and the use of the residual structure. According to the results in Table~\ref{tab5}, after integrating the convolutional branch, the mIoU increased by 0.12\%, indicating that the integration of the convolutional branch can further compensate for VMamba's shortcomings in local information extraction. When using the residual structure, the mIoU increased by 0.14\% without increasing computational complexity, suggesting that residual connections can optimize the feature extraction process, reduce information loss, and allow the network to better extract features from each layer. When both the convolutional branch and the residual structure are used together, the mIoU increased by 0.29\%. Considering both accuracy and computational complexity, we conclude that the combination of the two is the optimal choice.

\subsubsection{Effect of MFMSBlock}
To explore the effect of introducing multi-frequency information on enhancing information integrity and the role of Adaptive 1D Conv in improving information transmission precision, we conducted ablation experiments on these two key modules for comparative analysis. As shown in Table~\ref{tab6}, the introduction of multi-frequency information further improves the model's segmentation accuracy, which fully demonstrates that incorporating multi-frequency information enhances information integrity and better captures complex information. Additionally, to prove that the local cross-channel interaction strategy in the MFMSBlock, achieved through Adaptive 1D Conv without dimensionality reduction, can effectively avoid the information loss caused by the dimensionality reduction in fully connected layers and ensure the precision of information transmission. We compared the use of fully connected layers and Adaptive 1D Conv to calculate channel weights for multi-frequency aggregated features. The results show that Adaptive 1D Conv improves mIoU by 0.26\% compared to using fully connected layers (FC), fully demonstrating the effectiveness of Adaptive 1D Conv.
\begin{table*}[t]
\belowrulesep=3pt
\aboverulesep=3pt   
\setlength{\tabcolsep}{6pt}
\caption{\textbf{Ablation results for each module of the MFMSBlock.}}
\centering
\begin{tabular}{cccc|ccc}
\hline
Multi-Scale &Multi-Frequency &Adaptive 1D Conv &FC &mIoU(\%) &FLOPs(Gbps) &Params(Mb)\\
\hline
\checkmark &$\times$ &$\times$ &\checkmark &75.63 &5.69 &31.24\\
\checkmark &\checkmark &$\times$ &\checkmark &75.71 &5.71 &30.94\\
\checkmark &\checkmark &\checkmark &$\times$ &75.97 &5.71 &30.84\\
\hline
\end{tabular}
\label{tab6}
\end{table*}

\begin{table}[H]
\belowrulesep=2pt
\aboverulesep=2pt   
\setlength{\tabcolsep}{6pt}
\caption{\textbf{Comparison of computational complexity between different methods.}}
\centering
\begin{tabular}{c|ccc}
\hline
Method &FLOPs(Gbps) &Params(Mb) &mIoU(\%)\\
\hline
BANet\cite{b25} &3.26 &\underline{12.72} &64.13\\
ABCNet\cite{b26} &3.91 &13.39 &71.85\\
UNetFormer\cite{b7} &\textbf{2.94} &\textbf{11.68} &70.63\\
${\rm A^{2}}$-FPN\cite{b27} &10.46 &22.82 &72.55\\
MAResU-Net\cite{b28} &8.78 &23.27 &74.05\\
MANet\cite{b14} &19.45 &35.86 &74.11\\
CMTFNet\cite{b29} &8.57 &30.07 &\underline{75.05}\\
Rs3Mamba\cite{b11} &15.82 &49.66 &73.18\\
CM-UNet\cite{b12} &\underline{3.17} &13.55 &74.10 \\
CHVM-UNet(ours) &5.71 &30.84 &\textbf{75.97}\\
\hline
\end{tabular}
\label{tab7}
\end{table} 

\subsection{Model Complexities}
We tested the computational complexity of all models using random tensor data with a size of $3\times256\times256$. The computational complexity and parameter scale of each method are compared in Table~\ref{tab7}, all assessed under equivalent environmental settings. As shown in Table~\ref{tab7}, While CVMH-UNet requires more computational resources than streamlined networks(such as ABCNet), its segmentation accuracy is significantly higher than that of these lightweight models. In contrast to other networks founded on CNN or Transformer principles (such as CMTFNet) and networks employing complex fusion strategies (such as MANet), CVMH-UNet exhibits a clear advantage in computational complexity. In particular, when compared to other VMamba-based networks such as Rs3Mamba and CM-UNet, CVMH-UNet achieves the highest segmentation accuracy. In summary, the results of the experiments show that the proposed CVMH-UNet effectively reduces computational costs while maintaining high segmentation accuracy, successfully balancing segmentation accuracy and computational efficiency.

\section{Conculusion}\label{s6}
This study presents CVMH-UNet, designed for segmenting high-resolution remote sensing images semantically. In CVMH-UNet, the CVSSBlock is designed as  the fundamental component of both the encoder and decoder. This module achieves in-depth exploration of both global and local features by introducing the CS2D scanning method in VMamba and integrating convolutional branches, thereby improving model accuracy and efficiency. At the same time, MFMSBlock is used to replace traditional skip connections to achieve refined feature fusion. MFMSBlock enhances information integrity and transmission precision during the feature fusion process by introducing multi-frequency information and Adaptive 1D Conv, and it provides additional local detail information at different scales through point-wise convolution branches to mitigate inconsistencies between features of different scales. Experimental results on classic high-resolution remote sensing datasets demonstrate that, as matched against existing segmentation methods, the proposed CVMH-UNet achieves optimal performance in segmentation accuracy and strikes a good equilibrium between segmentation precision and computational complexity. Additionally, ablation studies verify the potential of CS2D in CVSSBlock for efficiently extracting global features, the effectiveness of integrated convolutional branches in improving the local feature extraction capabilities of Vision Mamba, and the benefits of residual connections in optimizing feature extraction and reducing information loss. The experiments also demonstrate the role of multi-frequency information in MFMSBlock for enhancing information integrity and the contribution of Adaptive 1D Conv in improving information transmission precision. These results collectively validate the effectiveness of the proposed modules. Future work will involve additional optimization of the network structure, along with ongoing exploration of the potential and applications of Vision Mamba in remote sensing.


\begin{thebibliography}{1}

  \bibitem{b1}
  Liu, Yongcheng, et al.``Semantic labeling in very high resolution images via a self-cascaded convolutional neural network." \textit{ISPRS journal of photogrammetry and remote sensing}, vol.145, pp.78-95, 2018.

  \bibitem{b2}
  Yao, Huang, Rongjun Qin, and Xiaoyu Chen. "Unmanned aerial vehicle for remote sensing applications—A review." \textit{Remote Sensing}, vol.11, no.12, pp.1443, 2019.

  \bibitem{b3}
  Yan, Liang, et al. "Triplet adversarial domain adaptation for pixel-level classification of VHR remote sensing images." \textit{IEEE Transactions on Geoscience and Remote Sensing}, vol.58, no.5, pp.3558-3573, 2019.

  \bibitem{b4}
  Li, Rui, et al. "Land cover classification from remote sensing images based on multi-scale fully convolutional network."  \textit{Geo-spatial information science}, vol.25, no.2, pp.278-294, 2019.

  \bibitem{b5}
  Liu, Huan, et al. "Central attention network for hyperspectral imagery classification." \textit{IEEE Transactions on Neural Networks and Learning Systems}, vol.34, no.11, pp.8989-9003, 2022.

  \bibitem{b6}
  Zhang, Mengmeng, et al. "Hyperspectral and LiDAR data classification based on structural optimization transmission." \textit{IEEE Transactions on Cybernetics}, vol.53, no.5, pp.3153-3164, 2022.

  \bibitem{b7}
  Zhang, Yuxiang, et al. "Single-source domain expansion network for cross-scene hyperspectral image classification." \textit{IEEE Transactions on Image Processing}, vol.32, pp.1498-1512, 2023.

  \bibitem{b8}
  Lian, Renbao, et al. "Road extraction methods in high-resolution remote sensing images: A comprehensive review." \textit{IEEE Journal of Selected Topics in Applied Earth Observations and Remote Sensing}, vol.13, pp.5489-5507, 2020.
 
  \bibitem{b9}
  Y. Guo, X. Jia and D. Paull, ``Effective Sequential Classifier Training for SVM-Based Multitemporal Remote Sensing Image Classification," in \textit{IEEE Transactions on Image Processing}, vol. 27, no. 6, pp. 3036-3048, Jun. 2018. 
  
  \bibitem{b10}
  Pal, Mahesh, ``Random forest classifier for remote sensing classification." \textit{International journal of remote sensing} vol.26, no.1, pp.217-222, 2005.

  \bibitem{b11}
  P. Kr\"ahenb\"uhl and V. Koltun, ``Efficient inference in fully connected CRFs with Gaussian edge potentials," in \textit{Proc. Adv. Neural Inf. Process. Syst. (NeurIPS)}, pp. 109-117, 2011.

  \bibitem{b12}
  O. Ronneberger, P. Fischer, and T. Brox, ``U-Net: Convolutional networks for biomedical image segmentation," in \textit{Proc. Int. Conf. Med. Image Comput. Comput.-Assist. Intervent.}, Cham, Switzerland: Springer, pp. 234-241, 2015.
  
  \bibitem{b13}
  Dosovitskiy, Alexey. "An image is worth 16x16 words: Transformers for image recognition at scale.", \textit{arXiv preprint arXiv:2010.11929}, 2020.

  \bibitem{b14}
  Zheng, Sixiao, et al. "Rethinking semantic segmentation from a sequence-to-sequence perspective with transformers.", \textit{Proceedings of the IEEE/CVF conference on computer vision and pattern recognition}, pp.6881-6890, 2021.

  \bibitem{b15}
  Wang, Wenxiao, et al. "Crossformer++: A versatile vision transformer hinging on cross-scale attention." \textit{IEEE Transactions on Pattern Analysis and Machine Intelligence}, 2023.

  \bibitem{b16}
  He, Xin, et al. "Swin transformer embedding UNet for remote sensing image semantic segmentation." \textit{IEEE Transactions on Geoscience and Remote Sensing}, vol.60, pp.1-15, 2022.

  \bibitem{b17}
  Cao, Hu, et al. ``Swin-unet: Unet-like pure transformer for medical image segmentation." \textit{European conference on computer vision}, Cham: Springer Nature Switzerland, pp.205-218, 2022.

  \bibitem{b18}
  Z. Liu et al. ``Swin Transformer: Hierarchical Vision Transformer using Shifted Windows,"  \textit{2021 IEEE/CVF International Conference on Computer Vision (ICCV)}, Montreal, QC, Canada, pp. 9992-10002, 2021.

  \bibitem{b19}
  Wang, Libo, et al. ``UNetFormer: A UNet-like transformer for efficient semantic segmentation of remote sensing urban scene imagery." \textit{ISPRS Journal of Photogrammetry and Remote Sensing} vol.190, pp.196-214, 2022.

  \bibitem{b20}
  Gu, Albert, and Tri Dao. ``Mamba: Linear-time sequence modeling with selective state spaces." \textit{arXiv preprint arXiv:2312.00752}, 2023.

  \bibitem{b21}
  Zhu, Lianghui, et al. ``Vision mamba: Efficient visual representation learning with bidirectional state space model." \textit{arXiv preprint arXiv:2401.09417}, 2024.

  \bibitem{b22}
  Y. Liu, Y. Tian, Y. Zhao, H. Yu, L. Xie, Y. Wang, Q. Ye, and Y. Liu. ``Vmamba: Visual state space model."  \textit{arXiv preprint arXiv:2401.10166}, 2024.

  \bibitem{b23}
  Ma, Xianping, Xiaokang Zhang, and Man-On Pun. ``RS 3 Mamba: Visual State Space Model for Remote Sensing Image Semantic Segmentation." \textit{IEEE Geoscience and Remote Sensing Letters}, 2024.

  \bibitem{b24}
  Liu, Mushui, et al. ``CM-UNet: Hybrid CNN-Mamba UNet for Remote Sensing Image Semantic Segmentation." \textit{arXiv preprint arXiv:2405.10530}, 2024.

  \bibitem{b25}
  Dai, Yimian, et al. ``Attentional feature fusion." \textit{Proceedings of the IEEE/CVF winter conference on applications of computer vision.}, pp.3560-3569, 2021.

  \bibitem{b26}
  Li, Rui, et al. ``Multiattention network for semantic segmentation of fine-resolution remote sensing images." \textit{IEEE Transactions on Geoscience and Remote Sensing}, vol.60 , pp.1-13, 2021.

  \bibitem{b27}
  Wang, Qilong, et al. ``ECA-Net: Efficient channel attention for deep convolutional neural networks." \textit{Proceedings of the IEEE/CVF conference on computer vision and pattern recognition.}, pp.11534-11542, 2020.

  \bibitem{b28}
  Qin, Zequn, et al. ``Fcanet: Frequency channel attention networks." \textit{Proceedings of the IEEE/CVF international conference on computer vision.}, pp.783-792, 2021.

  \bibitem{b29}
  Gu, Albert, Karan Goel, and Christopher Ré. ``Efficiently modeling long sequences with structured state spaces." \textit{arXiv preprint arXiv:2111.00396}, 2021.

  \bibitem{b30}
  Huang, Tao, et al. ``Localmamba: Visual state space model with windowed selective scan." \textit{arXiv preprint arXiv:2403.09338}, 2024.

  \bibitem{b31}
  Li, Yapeng, et al. ``MambaHSI: Spatial-Spectral Mamba for Hyperspectral Image Classification." \textit{IEEE Transactions on Geoscience and Remote Sensing}, 2024.

  \bibitem{b32}
  Hu, Jie, Li Shen, and Gang Sun. ``Squeeze-and-excitation networks." \textit{Proceedings of the IEEE conference on computer vision and pattern recognition.}, pp.7132-7141, 2018.

  \bibitem{b33}
  Woo, Sanghyun, et al. ``Cbam: Convolutional block attention module." \textit{Proceedings of the European conference on computer vision (ECCV).}, pp.3-19, 2018.

  \bibitem{b34}
  Roy, Abhijit Guha, Nassir Navab, and Christian Wachinger. ``Recalibrating fully convolutional networks with spatial and channel “squeeze and excitation” blocks." \textit{IEEE transactions on medical imaging}, vol.38, no.2, pp.540-549, 2018.

  \bibitem{b35}
  Lee, HyunJae, Hyo-Eun Kim, and Hyeonseob Nam. ``Srm: A style-based recalibration module for convolutional neural networks." \textit{Proceedings of the IEEE/CVF International conference on computer vision.}, pp.1854-1862, 2019.

  \bibitem{b36}
  Li, Xiang, et al. ``Selective kernel networks." \textit{Proceedings of the IEEE/CVF conference on computer vision and pattern recognition.}, pp.510-519, 2019.

  \bibitem{b37}
  Wang, Libo, et al. ``Transformer meets convolution: A bilateral awareness network for semantic segmentation of very fine resolution urban scene images." \textit{Remote Sensing}, vol.13, no.16, pp.3065  13.16, 2021.

  \bibitem{b38}
  Li, Rui, et al. ``ABCNet: Attentive bilateral contextual network for efficient semantic segmentation of Fine-Resolution remotely sensed imagery." \textit{ISPRS journal of photogrammetry and remote sensing}, vol.181, pp.84-98, 2021.

  \bibitem{b39}
  Li, Rui, et al. ``A2-FPN for semantic segmentation of fine-resolution remotely sensed images."  \textit{International journal of remote sensing}, vol.43, no.3, pp.1131-1155, 2022.

  \bibitem{b40}
  Li, Rui, et al. ``Multistage attention ResU-Net for semantic segmentation of fine-resolution remote sensing images." \textit{IEEE Geoscience and Remote Sensing Letters}, vol.19, pp.1-5, 2021.

  \bibitem{b41}
  Wu, Honglin, et al. ``CMTFNet: CNN and multiscale transformer fusion network for remote sensing image semantic segmentation." \textit{IEEE Transactions on Geoscience and Remote Sensing}, 2023.

  
  
  
  
  
  
  
\end{thebibliography}
\end{document}